\documentclass[a4paper, 12pt]{article}

\usepackage{graphicx}        % standard LaTeX graphics tool
\usepackage{amsmath}
\usepackage{amssymb}
\begin{document}

\title{Some issues in robust clustering}
% \titlerunning{Short Title} %for an abbreviated version of your contribution title if the original one is too long
\author{Christian Hennig\\
\small
Dipartimento di Scienze Statistiche ``Paolo Fortunati'',\\ 
\small
University of Bologna,\\ 
\small Via delle Belle Arti 41, 40126 Bologna, Italy,\\
\small christian.hennig@unibo.it}
%\authorrunning{Christian Hennig} %If there are more than two authors, please, abbreviate the authors' list using 'et al'

%
% Use the package "url.sty" to avoid
% problems with special characters
% used in your e-mail or web address
%
\maketitle

\begin{abstract}
Some key issues in robust clustering are discussed with focus on Gaussian mixture model based clustering, namely the formal definition of outliers, ambiguity between groups of outliers and clusters, the interaction between robust clustering and the estimation of the number of clusters, the essential dependence of (not only) robust clustering on tuning decisions, and shortcomings of existing measurements of cluster stability when it comes to outliers.  

{\bf Keywords:} Gaussian mixture model, trimming, noise component, number of clusters, user tuning, cluster stability\\
MSC2010 classification: 62H30
\end{abstract}

\section{Introduction}
\label{sec:1}
This is accepted for publication in {\it P. Brito, J. G. Dias, B. Lausen, A. Montanari, R. Nugent (eds.) Classification and Data Science in the Digital Age (Proeedings of IFCS-2022 Porto), Springer (2023).}
~\\~\\
Cluster analysis is about finding groups in data. Robust statistics is about methods that are not affected strongly by deviations from the statistical model assumptions or moderate changes in a data set. Particular attention has been paid in the robustness literature to the effect of outliers. Outliers and other model deviations can have a strong effect on cluster analysis methods as well. There is now much work on robust cluster analysis, see \cite{BanDav12,Ritter15,GGMMH16} for overviews.

There are standard techniques of assessing robustness such as the influence function and the breakdown point \cite{HubRon09} as well as simulations involving outliers, and these have been applied to robust clustering as well \cite{Ritter15,GGMMH16}.     

Here I will argue that due to the nature of the cluster analysis problem, there are issues with the standard reasoning regarding robustness and outliers.  

The starting point will be clustering based on the Gaussian mixture model, 
for details see \cite{BCMR19}. For this approach, $n$ observations are assumed i.i.d. 
with density
\begin{displaymath}
  f_\eta(x)=\sum_{k=1}^K \pi_k \varphi_{\mu_k,\Sigma_k}(x),
\end{displaymath}
$x\in\mathbb{R}^p$, with $K$ mixture components with proportions $\pi_k$, $\varphi_{\mu_k,\Sigma_k}$ being the Gaussian density with mean
vectors $\mu_k$, covariance matrices $\Sigma_k,\ k=1,\ldots,K$, $\eta$ being a vector of all parameters. For given $K$, $\eta$ can be estimated by maximum likelihood (ML) using the EM-algorithm, as implemented for example in the R-package ``mclust''. A standard approach to estimate $K$ is the optimisation of the Bayesian Information Criterion (BIC). Normally, mixture components are interpreted as clusters, and observations $x_i,\ i=1,\ldots,n$, can be assigned to clusters using the estimated posterior probability that $x_i$ was generated by mixture component $k$. 
A problem with ML estimation is that the likelihood degenerates if all observations assigned to a mixture component lie on a lower dimensional hyperplane, i.e, a $\Sigma_k$ has an eigenvalue of zero. This can be avoided by placing constraints on the eigenvalues of the covariance matrices \cite{GGGIM18}. Alternatively, a non-degenerate local optimum of the likelihood can be used, and if this cannot be found, constrained covariance matrix models (such as $\Sigma_1=\ldots=\Sigma_K$) can be fitted instead, as is the default of mclust.  Several issues with robustness that occur here are also relevant for other clustering approaches. 
 
\section{Outliers vs. clusters}
It is well known that the sample mean and sample covariance matrix as estimators of the parameters of a single Gaussian distribution can be driven to breakdown by a single outlier \cite{HubRon09}. Under a Gaussian mixture model with fixed $K$, an outlier must be assigned to a mixture component $k$ and will break down the estimators of $\mu_k, \Sigma_k$ (which are weighted sample means and covariance matrices) for that component in the same manner; the same holds for a cluster mean in $k$-means clustering.

Addressing this issue, and dealing with more outliers in order to achieve a high breakdown point, is a starting point for robust clustering. Central ideas are trimming a proportion of observations \cite{GarGor99}, adding a ``noise component'' with constant density to catch the outliers \cite{CorHen17,BCMR19}, mixtures with more robust component-wise estimators such as mixtures of heavy-tailed distributions (Sec. 7 of \cite{McLPee00}). 
  
But cluster analysis is essentially different from estimating a homogeneous population. Given a data set with $K$ clear Gaussian clusters and standard ML-clustering, consider adding a single outlier that is far enough away from the clusters. Assuming a lower bound on covariance matrix eigenvalues, the outlier will form a one-point cluster, the mean of which will diverge with the added outlier, and the original clusters will be merged to form $K-1$ clusters \cite{Hennig04}. 

The same will happen with a group of several outliers being close together, once more added far enough away from the Gaussian clusters. ``Breakdown'' of an estimator it is usually understood as the estimator becoming useless. It is questionable that this is the case here. In fact, the ``group of outliers'' can well be interpreted as a cluster in its own right, and putting all these points together in a cluster could be seen as desirable behaviour of the ML estimator, at least if two of the original $K$ clusters are close enough to each other that merging them will produce a cluster that is fairly well fitted by a single Gaussian distribution; note that the Gaussian mixture model does not assume strong separation between components, and a mixture of two Gaussians may be unimodal and in fact very similar to a single Gaussian. 
A breakdown point larger than a given $\alpha,\ 0<\alpha<\frac{1}{2}$ may not be seen as desirable in cluster analysis
if there can be clusters containing a proportion of less than $\alpha$ of the data, as a larger breakdown point will 
stop a method from taking such clusters (when added in large distance from the rest of the data) appropriately into account.

The core problem is that it is not clear what distinguishes a group of outliers from a legitimate cluster. I am not aware of any formal definition of outliers and clusters in the literature that allows this distinction. Even a one-point cluster is not necessarily invalid. Here are some possible and potentially conflicting aspects of such a distinction.
\begin{itemize}
\item A certain minimum size may be required for a cluster; smaller groups of points may be called outliers.
\item Groups of points in low density areas of the data may be called outliers. Note that this particularly means that very widely spread Gaussian mixture components would also be defined as outliers, deviating from the standard interpretation of Gaussian mixture components as clusters.
\item Members of non-Gaussian mixture components may be called outliers. This does not seem to be a good idea, because Gaussianity cannot be assessed for too small groups of observations, and furthermore in practice model assumptions are never perfectly fulfilled, and it may be desirable to interpret homogeneous or unimodal non-Gaussian parts of the data as ``cluster'' and fit them by a Gaussian component.
\item The term ``outlier'' suggests that outliers lie far away from most other observations, so it may be required that outliers are farther away from the clusters than the clusters are from each other. But this would be in conflict with the intuition that strong separation is usually seen as a desirable feature for well interpretable clusters. It may only be reasonable in applications in which there is prior information that there is limited variation even between clusters, as is implied by certain Bayesian approaches to clustering \cite{MaFrGr17}. 
\item The term ``cluster'' may be seen as flexible enough that a definition of an outlier is not required. Clustering should accommodate whatever is ``outlying'' by fitting it by one or more further clusters, if necessary of size one (single linkage clustering can be useful for outlier detection, even though it is inappropriate for most clustering problems).  
\end{itemize}
Most of these items require specific decisions that cannot be made in any objective and general manner, 
but only taking into account subject matter information, such as the minimum size of valid clusters or the 
density level below which observations are seen as outliers (potentially compared to density peaks in the distribution). This
implies that an appropriate treatment of outliers in cluster analysis cannot be expected to be possible without user tuning.

\section{Robustness and the number of clusters} 
The last item suggests that there is an interplay between outlier identification and the number of clusters, and that adding clusters might be a way of dealing with outliers; as long as clusters are assumed to be Gaussian, a single additional component may not be enough. More generally, concentrating robustness research on the case of fixed $K$ may be seen as unrealistic, because $K$ is rarely known, although estimating $K$ is a notoriously difficult problem even without worrying about outliers \cite{Hennig16}.

The classical robustness concepts, breakdown point and influence function, assume parameters from $\mathbb{R}^q$ with fixed $q$. If $K$ is not fixed, the number of parameters is not fixed either, and the classical concepts do not apply. 

As an alternative to the breakdown point, \cite{Hennig08} defined a ``dissolution point''. Dissolution is measured in terms of cluster memberships of points rather than in terms of parameters, and is therefore also applicable to nonparametric clustering methods. Furthermore, dissolution applies to individual clusters in a clustering; certain clusters may dissolve, i.e., there may be no sufficiently similar cluster in a new clustering computed after, e.g., adding an outlier; and others may not dissolve. This does not require $K$ to be fixed; the definition is chosen so that if a clustering changes from $K$ to $L<K$ clusters, at least $K-L$ clusters dissolve.

\cite{Hennig04,Hennig08} showed that when estimating $K$ using the BIC and standard ML estimation, reasonably well separated clusters do not dissolve when adding possibly even a large percentage of outliers (this does not hold for every method to estimate the number of clusters, see \cite{Hennig08}). Furthermore, \cite{Hennig08} showed that no method with fixed $K$ can be robust for data in which $K$ is misspecified - already \cite{GarGor99} had found that robustness features in clustering generally depend on the data.    
  
An implication of these results is that even in the fixed $K$ problem, the standard ML method can be a valid competitor regarding robustness if it comes with a rule that allows to add one or possibly more clusters that can then be used to fit the outliers (this is rarely explored in the literature, but \cite{McLPee00}, Sec. 7.7, show an example in which adding a single component does not work very well).  

An issue with adding clusters to accommodate outliers is that in many applications it is appropriate to distinguish between
meaningful clusters, and observations that cannot be assigned to such clusters (often referred to as ``noise'').
Even though adding clusters of outliers can formally prevent the dissolution of existing clusters, it may be misleading to interpret
the resulting clusters as meaningful, and a classification as outliers or noise can be more useful. This is provided
by the trimming and noise component approaches to robust clustering. Also some other clustering methods such as the density-based
DBSCAN \cite{EKSX96} provide such a distinction. On the other hand, modelling clusters by heavy-tailed distributions such as in mixtures of t-distributions
will implicitly assign outlying observations to clusters that potentially are quite far away. For this reason, \cite{McLPee00}, Sec, 7.7, provide an additional outlier identification rule on top of the mixture fit. \cite{FarPun20} even distinguish between ``mild'' outliers that are modelled as having a larger variance around the same mean, and ``gross'' outliers to be trimmed.
The variety of approaches can be connected  to the different meanings that outliers can have in applications. They can be erroneous, they can be irrelevant noise, but they can also be caused by unobserved but relevant special conditions (and would as such qualify as meaningful clusters), or they could be valid observations legitimately belonging to a meaningful cluster that regularly produces observations further away from the centre than modelled by a Gaussian distribution.

Even though currently there is no formal robustness property that requires both the estimation of $K$ and an identification or downweighting of outliers,  
there is demand for a method that can do both. 
% The previous discussion implies that this cannot be done without tuning decisions. 

Estimating $K$ comes with an additional difficulty that is relevant in connection
with robustness. As mentioned before, in clustering based on the 
Gaussian mixture model normally every mixture component will be interpreted as
a cluster. In reality, however, meaningful clusters are not perfectly Gaussian.
Gaussian mixtures are very flexible for approximating non-Gaussian distributions.
Using a consistent method for estimating $K$ means that for large enough $n$
a non-Gaussian cluster will be approximated by several Gaussian mixture
components. The estimated $K$ will be fine for producing a Gaussian mixture 
density that fits the data well, but it will overestimate the number of 
interpretable clusters. The estimation of $K$, if interpreted as the number 
of clusters, relies on precise Gaussianity of the clusters, and is as such
itself riddled with a robustness problem; in fact slightly non-Gaussian clusters
may even drive the estimated $K\to\infty$ if $n\to\infty$ 
\cite{Hennig10,HenCor21}. 

This is connected with the more fundamental problem that there is no unique
definition of a cluster either. The cluster analysis 
user needs to specify the cluster concept of interest even before robustness
considerations, and arguably different clustering methods imply different
cluster concepts \cite{Hennig16}. A Gaussian mixture model defines clusters 
by the Gaussian distributional shape 
(unless mixture components are merged to form clusters \cite{Hennig10}). 
Although this can be motivated in some real situations, robustness 
considerations require that distributional
shapes fairly close to the Gaussian should be accepted as clusters
as well, but this requires another specification, namely 
how far from a Gaussian a cluster is allowed to be, or alternatively how 
separated Gaussian components have to be in order to count as separated 
clusters. A similar problem can also
occur in nonparametric clustering; if clusters are associated with density 
modes or level sets, the cluster concept depends on how weak a mode or gap 
between high level density sets is allowed to be to be treated as meaningful.

\cite{HenCor21} propose a parametric bootstrap approach to simultaneously
estimate $K$ and assign outliers to a noise component. 
This requires two basic
tuning decisions. The first one regards the minimum 
percentage of observations so that a researcher is willing to add another
cluster if the noise component can be reduced by this amount. The second one
specifies a tolerance that allows a data subset to count as a cluster even 
though it deviates to some extent from what is expected under a perfectly
Gaussian distribution. There is a third tuning parameter that is in effect
for fixed $K$ and tunes how much of the tails of a non-Gaussian cluster can 
be assigned to the noise in order to improve the Gaussian appearance of the
cluster. One could even see the required constraints on covariance matrix
eigenvalues as a further tuning decision. Default values can be provided, but
situations in which matters can be improved deviating from default values are
easy to construct.

\section{More on user tuning}
User tuning is not popular, as it is often difficult to make appropriate tuning decisions. Many scientists believe that subjective user decisions threaten scientific objectivity, and also background knowledge dependent choices cannot be 
made when investigating a method's performance by theory and simulations. 
The reason why user tuning is indispensable in robust cluster analysis is 
that it is required in order to make the problem well defined. The distinction 
between clusters and outliers is an interpretative one that no automatic method
can make based on the data alone. Regarding the number of clusters, imagine two
well separated clusters (according to whatever cluster concept of interest), and
then imagine them to be moved closer and closer together. Below what distance 
are they to be considered a single cluster? This is essentially a tuning 
decision that the data cannot make on their own. 

There are methods that do not
require user tuning. Consider the mclust implementation of Gaussian mixture
model based clustering. The number of clusters is by default estimated by the 
BIC. As seen above, this is not really appropriate for large data sets, but
its derivation is essentially asymptotic, so that there is no theoretical 
justification for it for small data sets either. Empirically it often but not always works well, and there is little investigation of whether it tends to make the ``right'' decision in ambiguous situations where it is not clear without user tuning what it even means to be ``right''. Covariance matrix constraints in 
mclust are not governed by a tuning of eigenvalues or their ratios to be 
specified by the user. Rather the BIC decides between different covariance matrix models, but this can be erratic and unstable, as it depends on whether the EM-algorithm gets caught in a degenerate likelihood maximum or not, and in situations where two or more covariance matrix models have similar BIC values (which happens quite often), a tiny change in the data can result in a different covariance matrix model being selected, and substantial changes in the clustering. 
A tunable eigenvalue condition can result in much smoother behaviour. When it
comes to outlier identification, mclust offers the addition of a uniform ``noise'' mixture component governed by the range of the data,
again supposedly without user tuning. This starts from an initial noise estimation that requires tuning (Sec. 3.1.2 of \cite{BCMR19}) and is less robust in terms of breakdown and dissolution than trimming and the improper noise component, both of which require tuning \cite{Hennig04,Hennig08}. The ICL, an alternative to the BIC (Sec. 2.6 of \cite{BCMR19}), on the other hand, is known to merge different Gaussian mixture components already at a distance at which they intuitively still seem to be separated clusters. 
Similar comments apply to the mixture of t-distributions; it
requires user tuning for identifying outliers, scatter matrix constraints, 
and it has the same issues with BIC and ICL as the Gaussian mixture.

Summarising, both the identification of and robustness against outliers and the
estimation of the number of clusters require tuning in order to be well
defined problems; user tuning can only be avoided by taking tuning decisions
out of the user's hands and making them internally, which will work in 
some situations and fail in others, and the impression 
of automatic data driven decision making that a user may have is rather an 
illusion. This, however, does not free method designers from the necessity to
provide default tunings for experimentation and cases in which the users do
not feel able to make the decisions themselves, and tuning guidance
for situations in which more information is available. A decision regarding
the smallest valid size of a cluster is rather well interpretable; a decision
regarding admissible covariance matrix eigenvalues is rather difficult and 
abstract. 
\section{Stability measurement}
Robustness is closely connected to stability.
Both experimental and theoretical investigation of the stability of clusterings
require formal stability measurements, usually comparing two clusterings
on the same data (potentially modified by replacing or adding observations). 
Not assuming any parametric model, proximity measures such
as the Adjusted Rand Index (ARI; \cite{HubAra85}), the Hamming distance (HD; \cite{BeLuPa06}), or the Jaccard distance between individual clusters \cite{Hennig08} can be used. Note that \cite{BeLuPa06}, standard reference on cluster stability in the machine learning community, state that stability and instability are caused in the first place by ambiguities in the cluster structure of the data, rather than by a method's robustness or lack of it. Although the outlier problem is ignored in that paper, it is true that cluster analysis can have other stability issues that are as serious as or worse than gross outliers.

To my knowledge, none of the measures currently in use allow for a special treatment of a set of outliers or noise; either these have to be ignored, or treated just as any other cluster. Both ARI and HD, comparing clusterings $\mathcal{C}_1$ and $\mathcal{C}_2$, consider pairs of observations $x_i, x_j$ and check whether those that are in the same cluster in $\mathcal{C}_1$ are also in the same cluster in $\mathcal{C}_2$. An appropriate treatment of noise sets $N_1\in\mathcal{C}_1, N_2\in\mathcal{C}_2$  would require that $x_i, x_j\in N_1$ are not just in the same cluster in $\mathcal{C}_2$ but rather in $N_2$, i.e., whereas the numberings of the regular clusters do not have to be matched (which is appropriate because 
cluster numbering is meaningless), $N_1$ has to be matched to $N_2$. 
Corresponding re-definitions of these proximities will be useful to robustness
studies.

\section{Conclusion}
Key practical implications of the above discussions are:
\begin{itemize}
\item Outliers can be treated as forming their own clusters, or be collected in outlier/noise or trimmed sets, or be integrated in clusters of non-outliers. Which of these is appropriate depends on the nature of outliers in a given application.
\item Methods that do not identify outliers but add clusters in order to accommodate them are valid competitors of robust clustering methods, as are nonparametric density-based methods.
\item Cluster analysis involving estimating the number of clusters and robustness require tuning in order to define the problem they are meant to solve well. Method developers need to provide sensible defaults, but also to guide the users regarding a meaningful interpretation of the tuning decisions.       
\end{itemize}

\end{document}